\pdfoutput=1

\documentclass[11pt]{article}

\usepackage{acl}

\usepackage{times}
\usepackage{latexsym}
\usepackage{multirow}
\usepackage{graphicx}
\usepackage{booktabs}
\usepackage{bbding}
\usepackage{stfloats}
\usepackage{caption}
\usepackage{subcaption}

\newcommand{\tabincell}[2]{\begin{tabular}{@{}#1@{}}#2\end{tabular}}

\usepackage[T1]{fontenc}

\usepackage[utf8]{inputenc}

\usepackage{microtype}

\title{Investigating Zero- and Few-shot Generalization in Fact Verification}

\author{Liangming Pan\textsuperscript{$\dagger$} \\
  University of California, Santa Barbara \\
  \texttt{liangmingpan@ucsb.edu} \And
  Yunxiang Zhang\textsuperscript{$\dagger$} \\
  University of Michigan, Ann Arbor \\
  \texttt{yunxiang@umich.edu} \AND
  Min-Yen Kan \\
  National University of Singapore \\
  \texttt{kanmy@comp.nus.edu.sg} \\}

\begin{document}
\maketitle
\begin{abstract}
We explore \textit{zero- and few-shot generalization for fact verification (FV)}, which aims to generalize the FV model trained on well-resourced domains (\textit{e.g.}, Wikipedia) to low-resourced domains that lack human annotations.  
To this end, we first construct a benchmark dataset collection which contains 11 FV datasets representing 6 domains. We conduct an empirical analysis of generalization across these FV datasets, finding that current models generalize poorly. Our analysis reveals that several factors affect generalization, including dataset size, length of evidence, and the type of claims. Finally, we show that two directions of work improve generalization: 1) incorporating domain knowledge via pretraining on specialized domains, and 2) automatically generating training data via claim generation. 

\let\thefootnote\relax\footnotetext{\textsuperscript{$\dagger$}Equal Contribution.}






\end{abstract}

\section{Introduction}

With a rise in deliberate disinformation, \textit{Fact Verification (FV)}
has become an important NLP application.
FV aims to verify given claims with the evidence retrieved from plain text. 
Rapid progress has been made by training large neural models~\cite{DBLP:conf/acl/ZhouHYLWLS19,DBLP:conf/acl/LiuXSL20,DBLP:conf/acl/ZhongXTXDZWY20} on the FEVER dataset~\cite{DBLP:conf/naacl/ThorneVCM18}, containing more than 100K human-crafted (evidence, claim) pairs based on Wikipedia. Fact verification is also needed in other domains, including news, social media, and scientific documents. This has spurred the creation of a large number of FV datasets, such as COVID-Fact~\cite{DBLP:conf/acl/SaakyanCM20},
SciFact~\cite{DBLP:conf/emnlp/WaddenLLWZCH20}, and Climate-FEVER~\cite{DBLP:journals/corr/abs-2012-00614}. 

However, considering that human annotation is time-consuming, costly, and often biased, it is difficult to collect reliable human-labeled data in every domain that demands fact verification. We need to investigate how to build a generalizable fact verification system that adapts to new domains with zero or few samples. Critically, how can we leverage valuable (evidence, claim, label) annotations from rich-resourced domains (\textit{e.g.}, Wikipedia) to aid fact verification in the low-resourced ones (\textit{e.g.}, scholarly documents, and social media)?
Although FV datasets have been recently created in different domains, little analysis has shown whether FV models generalize across them and to what extent existing datasets can be leveraged to improve performance in these new domains. 

In this paper, we bridge this gap by conducting a comprehensive investigation of \textit{zero- and few-shot generalization in fact verification}. By conducting a holistic study of FV datasets to date, we first carefully select 8 datasets that have artificial or natural claims, human-annotated evidence, and two or three-class labels for our study. We then standardize their data formats as (evidence, claim, label) pairs and create dataset variants with different granularity of evidence, which gives us a total of 11 datasets. 
We then conduct a thorough empirical study of generalization and transfer across these 11 datasets. We train models on a source dataset, and then evaluate their performance on a target dataset, either without any additional target training examples (\textit{zero-shot setting}) or with a few additional target examples (\textit{few-shot setting}). 

We find that RoBERTa-based FV models tend to overfit the particular training set, generalizing poorly to other datasets. Our in-depth analysis shows generalization is related to several key factors, including dataset size, length of evidence, and the claim type. In particular, we find that Wikipedia-based artificial claims (\textit{e.g.}, FEVER) generalize well to natural claims in real-world domains with the growth of dataset size, in contrast to prior work that criticized crowd-sourced claims as having strong annotation bias and being unrepresentative of real-world misinformation~\cite{DBLP:conf/emnlp/SchusterSYFSB19}. Our few-shot generalization experiment further shows that fine-tuning on a small amount of target training data can substantially improve performance. 


Armed with the above insights, we explore two ways to improve the generalization of fact verification models. 1) \textit{Domain-specific Pretraining}: initializing the FV model with language models pretrained on specialized domains, and 2) \textit{Data Augmentation}: automatically generating training data for the target domain. Results show that these methods can noticeably improve generalization but still leave unsolved challenges such as inflexibility, high cost, and label consistency. 

To the best of our knowledge, this is the first work to perform a thorough investigation of generalization and transfer in fact verification. We open-sourced our dataset collection and codes to support future research towards a universal and robust fact verification system\footnote{\url{https://github.com/teacherpeterpan/Fact-Checking-Generalization}}. 


\section{Dataset Curation}

In this section, we describe the 11 fact verification datasets included in our study. 
We first describe the criteria for dataset selection (\S~\ref{sec:data_selection}), and then we introduce the dataset processing (\S~\ref{sec:data_processing}). We show the key characteristics of the datasets in Table~\ref{tbl:datasets_statistics}. 

\subsection{Dataset Selection}
\label{sec:data_selection}


A large number of datasets have recently been introduced to study various tasks for fact-checking, \textit{e.g.}, claim detection, evidence retrieval, fact verification, justification production, etc. Our focus, fact verification, in particular, takes a textual \textit{claim} and a piece of \textit{evidence} as input to predict the \textit{label} for the claim.  Let's define these aspects:

\vspace{0.15cm}

\noindent $\bullet$ \textbf{Claim}: Claims for fact verification are often textual, sentence-level statements, which are categorized into: 1) \textit{real-world natural claims} crawled from dedicated websites, textbooks, forums, etc. 2) \textit{artificial claims} written by crowd-workers. 

\vspace{0.1cm}

\noindent $\bullet$ \textbf{Evidence}: Evidence is the relevant information source for validating the claim. Textual sources, such as news articles, academic papers, and Wikipedia documents, are one of the most commonly used types of evidence. Based on the granularity, we categorize the evidence in existing datasets into: 1) \textit{document-level evidence} such as the Wikipedia page~\cite{DBLP:conf/naacl/ThorneVCM18}, news articles~\cite{DBLP:conf/conll/HanselowskiSSLG19}, and scientific papers~\cite{DBLP:conf/emnlp/WaddenLLWZCH20}. 2) \textit{sentence-level evidence} annotated by human experts in the relevant documents to support or refute each claim. 3) \textit{no evidence} is given for each claim; the model needs to retrieve evidence from a large knowledge source.  

\vspace{0.1cm}

\noindent $\bullet$ \textbf{Label}: The label definition for the claim also varies across datasets. The most common definition is the binary label with \texttt{supports} and \texttt{refutes}, and the three-class label, \textit{i.e.},  \texttt{supports/refutes/not enough info}. Some works~\cite{DBLP:conf/acl/Wang17,DBLP:conf/emnlp/AugensteinLWLHH19} also employ multi-class labels for more fine-grained degrees of truthfulness (\textit{e.g.} true, mostly-true, mixture, etc), where the number of labels vary greatly, ranging from 4 to 27. 


\begin{table*}[!t]
\centering
\resizebox{\textwidth}{!}{%
\renewcommand{\arraystretch}{1.1}
\begin{tabular}{ll|c|c|c|c|rr|rr}
\hline
  \multicolumn{2}{l|}{\multirow{2}{*}{Dataset}} &
  \multirow{2}{*}{Domain} &
  \multirow{2}{*}{Claim} &
  \multirow{2}{*}{Evidence} &
  \multirow{2}{*}{Label} &
  \multicolumn{2}{c|}{\# Claims} &
  \multicolumn{2}{c}{Avg. \# tokens} \\ \cline{7-10} 
 &  &  &  &  &  &  Train & Test & Claim & Evid. \\ \hline
 \multirow{4}{*}{\uppercase\expandafter{\romannumeral1}} & FEVER-sent &
  Wikipedia &
  artificial &
  sent-level &
  \multicolumn{1}{l|}{S (52\%), R (22\%), N (26\%)} &
  145,327 & 19,972 & 9.4 & 35.9 \\ 
& FEVER-para &
  Wikipedia &
  artificial &
  doc-level &
  \multicolumn{1}{l|}{S (52\%), R (22\%), N (26\%)} &
  145,327 & 19,972 & 9.4 & 368.7 \\ 
& VitaminC &
  Wikipedia &
  artificial &
  sent-level &
  \multicolumn{1}{l|}{S (50\%), R (35\%), N (15\%)} &
  370,653 & 63,054 & 12.6 & 29.5 \\
& FoolMeTwice &
  Wikipedia &
  artificial &
  sent-level &
  \multicolumn{1}{l|}{S (49\%), R (51\%)} &
  10,419 & 1,169 & 15.3 & 37.0\\ \hline
\multirow{7}{*}{\uppercase\expandafter{\romannumeral2}} & Climate-FEVER-sent &
  Climate &
  natural &
  sent-level &
  \multicolumn{1}{l|}{S (25\%), R (11\%), N (64\%)} &
  6,140 & 1,535 & 22.8 & 33.8 \\ 
& Climate-FEVER-para &
  Climate &
  natural &
  doc-level &
  \multicolumn{1}{l|}{S (47\%), R (19\%), N (34\%)} &
  1,103 & 278 & 22.9 & 168.9 \\ 
& Sci-Fact-sent &
  Science &
  natural &
  sent-level &
  \multicolumn{1}{l|}{S (43\%), R (22\%), N (35\%)} &
  868 & 321 & 13.8 & 61.9 \\ 
& Sci-Fact-para &
  Science &
  natural &
  doc-level &
  \multicolumn{1}{l|}{S (43\%), R (22\%), N (35\%)} &
  868 & 321 & 13.8 & 257.3 \\ 
& PubHealth &
  Health &
  natural &
  sent-level &
  \multicolumn{1}{l|}{S (60\%), R (36\%), N (4\%)} &
  8,370 & 1,050 & 15.7 & 137.6 \\ 
& COVID-Fact &
  Forum &
  natural &
  sent-level &
  \multicolumn{1}{l|}{S (32\%), R (68\%)} &
  3,268 & 818 & 12.4 & 82.5  \\ 
& FAVIQ &
  Question &
  natural &
  doc-level &
  \multicolumn{1}{l|}{S (50\%), R (50\%)} &
  17,008 & 4,260 & 15.2 & 304.9 \\ \hline
\end{tabular}%
}
\caption{List of the 11 fact verification datasets for our study and their characteristics.}  
\label{tbl:datasets_statistics}
\end{table*}

\paragraph{Selection Criteria.} 
We employ the following criteria to select the datasets for our study. 

\vspace{0.15cm}

\noindent $\bullet$ We consider both natural and artificial claims in various domains. 

\noindent $\bullet$ We consider the datasets with human-annotated document-level and sentence-level evidence. We exclude datasets without evidence or which provide only non-textual evidence; \textit{i.e.}, tables, knowledge bases, etc. 

\noindent $\bullet$ We only consider datasets with the binary or the three-class label annotation due to the difficulty of canonicalizing such multi-class labels. 


\vspace{0.15cm}

\noindent By conducting a holistic study of fact-checking datasets to date, eight different data sources meet our requirements. The full list of candidate datasets we investigate is given in Appendix \ref{sec:list-fv-datasets}. 

\subsection{Dataset Processing}
\label{sec:data_processing}

We then process the selected datasets as follows. 1) We convert each dataset to the unified format of claim-evidence-label triples ${(c_i, e_i, l_i)}_{i=1}^N$. The simplicity of this format allows us to focus on out-of-domain generalization, instead of other orthogonal challenges of fact-checking. 2) We create separate dataset variants by pairing each claim with the evidence in different granularity. This enables us to study the impact of evidence length on generalization. After processing, we obtain the final selection of 11 datasets used. We now briefly introduce the nature of each dataset and its specific processing. 


\paragraph{Group \uppercase\expandafter{\romannumeral1}: datasets with \textit{artificial} claims. }
These are based on Wikipedia articles and are often large in size. However, crowd-sourced claims are often written with minimal edits to reference sentences, leading to lexical biases such as the overuse of explicit negation~\cite{DBLP:conf/emnlp/SchusterSYFSB19}. 

\vspace{0.15cm}

\noindent $\bullet$ \textbf{FEVER}~\cite{DBLP:conf/naacl/ThorneVCM18} asks crowd-workers to mutate sentences from Wikipedia articles to create claims. 
We use the Wikipedia paragraph associated with each claim as its document-level evidence to construct the \textit{FEVER-para} dataset. We then use the sentence-level gold evidence for the \texttt{supports} and \texttt{refutes} claims to build the \textit{FEVER-sent} dataset. However, since sentence-level evidence is not available for \texttt{NEI} claims, we use the system of~\citet{DBLP:journals/corr/abs-1901-02534} to retrieve the evidence sentences, following~\citet{DBLP:conf/emnlp/AtanasovaWA20} and~\citet{DBLP:conf/acl/PanCXKW20}. 

\vspace{0.1cm}

\noindent $\bullet$ \textbf{VitaminC}~\cite{DBLP:conf/naacl/SchusterFB21} creates contrastive evidence pairs for each claim, in which evidence pairs are nearly identical in language and content, with the exception that one supports a claim while the other does not. 

\vspace{0.1cm}

\noindent $\bullet$ \textbf{FoolMeTwice}~\cite{DBLP:conf/naacl/EisenschlosDBBB21} designs a multi-player game that leads to diverse strategies for crafting claims (\textit{e.g.}, temporal inference) based on Wikipedia, resulting in more complex claims with less lexical overlap with the evidence.

\vspace{0.1cm}

\paragraph{Group \uppercase\expandafter{\romannumeral2}: datasets with \textit{natural} claims. }
These claims are collected from the Internet and then manually verified by professional fact checkers. They represent real-world claims, and originate from diverse domains, such as scholarly documents, news articles, forums, etc. However, due to the difficulty and high cost of manually verifying real-world claims, these datasets are limited in scale. 

\vspace{0.1cm}

\noindent $\bullet$ \textbf{Climate-FEVER}~\cite{DBLP:journals/corr/abs-2012-00614} consists of 1,535 real-life claims regarding climate-change collected from the Internet. The top five most relevant sentences from Wikipedia are retrieved as the evidence. Humans then annotate each sentence as supporting, refuting, or not enough information to validate the claim. We use the sentence-level annotation as the evidence for each claim to build the \textit{Climate-FEVER-sent}. We construct the document-level evidence for each claim by putting together all of its evidence sentences, which gives us the \textit{Climate-FEVER-para} version. 

\vspace{0.1cm}

\noindent $\bullet$ \textbf{Sci-Fact}~\cite{DBLP:conf/emnlp/WaddenLLWZCH20} consists of 1.4K expert-written scientific claims paired with evidence-containing abstracts annotated with labels and sentence-level rationale. We use the annotated rationale as the sentence-level evidence to build the \textit{Sci-Fact-sent}. We construct the \textit{Sci-Fact-para} version by using the evidence-containing abstract as the document-level evidence for each claim. 

\vspace{0.1cm}

\noindent $\bullet$ \textbf{PubHealth}~\cite{DBLP:conf/emnlp/KotonyaT20} contains 11.8K claims accompanied by journalist crafted, gold standard judgments to support/refute the claims. The claims are collected from five fact-checking websites, news headlines, and news reviews. We use the judgment texts as evidence to pair with each claim. 

\vspace{0.1cm}

\noindent $\bullet$ \textbf{COVID-Fact}~\cite{DBLP:conf/acl/SaakyanCM20} consists of 4,086 claims concerning the COVID19 pandemic crawled from the \textit{r/COVID19} subreddit. We use their sentence-level evidence annotated by crowd-workers as the evidence. 

\vspace{0.1cm}

\noindent $\bullet$ \textbf{FAVIQ}~\cite{DBLP:journals/corr/abs-2107-02153} contains 26k claims converted from natural ambiguous questions posed by real users. The answer-containing Wikipedia paragraph is provided as the document-level evidence for each claim. 

\vspace{0.15cm}

Many of the original datasets do not release their test set. Therefore, we use their original split of train/dev sets as our training and evaluation sets. We also standardize the naming of labels as \texttt{supports}, \texttt{refutes}, and \texttt{NEI}. We visualize the global structure of the datasets with tSNE ~\cite{JMLR:v9:vandermaaten08a} and analyze the domain divergence in Appendix \ref{sec:domain-divergence-analysis}. 


\section{Zero/Few-shot Generalization}

We now explore the generalization ability of fact verification models across the 11 datasets. We first formulate the task of zero/few-shot generalization. 

\paragraph{Task Formulation.}
Given a claim $\mathcal{C}$ and a piece of evidence $\mathcal{P}$ as inputs, a \textit{fact verification} model $\mathcal{F}$ predicts a label $\mathcal{Y}$ to verify whether $\mathcal{C}$ is supported, refuted, or can not be verified by the information in $\mathcal{P}$. In the \textit{zero-shot generalization} setting, we train models on one source FV dataset, and then evaluate its performance on a target test set, without any additional training data in the target dataset. In the \textit{few-shot generalization} setting, we assume we have a small amount of target training examples. 

\paragraph{Fact Verification Model.}
We use the RoBERTa-large~\cite{DBLP:journals/corr/abs-1907-11692} as the benchmark model for our study since it has achieved state-of-the-art results in many FV datasets. We concatenate the claim and evidence (\texttt{[CLS] \textit{claim} [SEP] \textit{evidence}}) and use it as input for a classification task to predict the label of the claim. We use the \texttt{roberta-large} (355M parameters) model provided by the HuggingFace library\footnote{\url{https://huggingface.co/roberta-large}}. 


\subsection{Zero-shot generalization results}
\label{sec:zero-shot-generalization}

Table~\ref{tbl:zero-shot-three-class-results} shows the zero-shot generalization results in macro-averaged F1 for the 3-class fact verification task on all the datasets that have \texttt{supports/refutes/NEI} labels, where we partition by dataset group: \textit{Group~I, top} (datasets with artificial claims); \textit{Group~II, bottom} (datasets with natural claims). 
%
In general, the RoBERTa model generalizes poorly in this zero-shot setup. Compared with the in-domain performance (training and testing on the same dataset), the best zero-shot generalization performance shows a large drop of 20.80\% on average. This shows that the FV model overfits to the particular dataset and generalizes poorly to unseen datasets. This validates prior work that shows the neural models are brittle when encountering out-of-distribution data. Taking a closer look, we further explore several research questions specific to fact verification behind this general trend. 

\begin{table*}[!t]
\setlength\tabcolsep{5pt}
\centering
\resizebox{0.9\textwidth}{!}{%
\begin{tabular}{l|ccc|ccccc}
\toprule
Train$\downarrow$ Test$\rightarrow$ & \tabincell{c}{FEVER\\-para} & \tabincell{c}{FEVER\\-sent} & VitaminC & \tabincell{c}{C-FEVER\\-para} & \tabincell{c}{C-FEVER\\-sent} & \tabincell{c}{SciFact\\-para} & \tabincell{c}{SciFact\\-sent} & PubHealth \\
\midrule
FEVER-para & --- & \textbf{72.81} & 43.87 & 20.83 & 40.90  & 22.09 & 28.10  & 9.05 \\
FEVER-sent & \textbf{55.57} & --- & \textbf{62.11} & 44.98 & \textbf{48.70} & \textbf{44.98} & \textbf{56.15} & 21.61 \\
VitaminC & 52.04 & 65.32 & --- & 42.32 & 44.40  & 44.14 & 50.55 & \textbf{21.97} \\
\midrule
C-FEVER-para & 17.86 & 20.04 & 10.59 & --- & 42.02 & 29.93 & 31.62 & 5.29 \\
C-FEVER-sent & 17.87 & 24.47 & 20.25 & \textbf{54.59} & --- & 25.84 & 39.39 & 8.47 \\
SciFact-para & 23.96 & 27.09 & 28.37 & 29.85 & 28.63 & --- & 44.68 & 6.78 \\
SciFact-sent & 16.86 & 24.50  & 29.22 & 20.61 & 32.50  & 29.00 & --- & 4.49 \\
PubHealth & 35.21 & 34.41 & 30.67 & 34.12 & 24.18 & 40.34 & 42.03 & --- \\
\midrule
SELF & 85.58 & 89.28 & 86.76 & 44.61 & 62.54 & 52.25 & 54.27 & 72.10 \\
\bottomrule
\end{tabular}%
}
\caption{Macro-F1 of \textbf{3-class fact verification} on the evaluation set for all datasets in a zero-shot generalization setup. Rows correspond to the training dataset and columns to the evaluated dataset. The row SELF corresponds to the in-domain performance (training and testing on the same target dataset). 
}
\label{tbl:zero-shot-three-class-results}
  \vspace{0.3cm}
\end{table*}

\begin{table*}[htbp]
\setlength\tabcolsep{2pt}
  \centering
\resizebox{\textwidth}{!}{%
\begin{tabular}{l|cccc|ccccccc}
\toprule
Train$\downarrow$ Test$\rightarrow$ & \tabincell{c}{FEVER\\-para} & \tabincell{c}{FEVER\\-sent} & VitaminC & \tabincell{c}{FoolMe\\Twice} & \tabincell{c}{C-FEVER\\-para} & \tabincell{c}{C-FEVER\\-sent} & \tabincell{c}{SciFact\\-para} & \tabincell{c}{SciFact\\-sent} & PubHealth & \tabincell{c}{COVID\\-Fact} & FAVIQ \\
\midrule
FEVER-para & --- & \textbf{94.91} & 71.56 & 72.50  & \textbf{77.40} & 76.04 & 72.29 & 75.92 & 44.75 & 56.82 & 55.09 \\
FEVER-sent & \textbf{89.08} & --- & \textbf{79.79} & 84.02 & 74.71 & \textbf{80.21} & \textbf{75.12} & \textbf{87.37} & 58.25 & 63.99 & 61.64 \\
VitaminC & 84.62 & 94.46 & --- & \textbf{84.57} & 62.80  & 54.59 & 62.37 & 69.31 & 55.32 & \textbf{70.32} & \textbf{62.98} \\
FoolMeTwice & 82.58 & 91.46 & 78.56 & --- & 71.38 & 78.24 & 69.22 & 84.19 & 56.81 & 58.68 & 59.23 \\
\midrule
C-FEVER-para & 33.37 & 33.72 & 52.56 & 34.15 & --- & 56.66 & 39.77 & 40.89 & 38.61 & 25.50  & 33.52 \\
C-FEVER-sent & 51.33 & 62.00 & 55.91 & 50.69 & 75.85 & --- & 66.72 & 72.08 & 55.54 & 43.04 & 36.53 \\
SciFact-para & 33.38 & 33.57 & 46.73 & 34.42 & 43.35 & 49.23 & --- & 41.27 & 38.69 & 26.64 & 33.69 \\
SciFact-sent & 33.40  & 33.64 & 36.91 & 33.77 & 42.63 & 42.46 & 44.02 & --- & 43.35 & 26.64 & 33.51 \\
PubHealth & 65.68 & 64.69 & 53.57 & 53.55 & 53.92 & 61.78 & 68.75 & 71.01 & --- & 40.95 & 50.89 \\
COVID-Fact & 70.94 & 76.16 & 37.22 & 63.02 & 44.13 & 51.71 & 63.60  & 76.29 & \textbf{60.06} & --- & 46.93 \\
FAVIQ & 74.57 & 73.80  & 59.14 & 59.67 & 64.92 & 60.49 & 59.08 & 52.64 & 40.15 & 50.25 & --- \\
\bottomrule
\end{tabular}%
}
\caption{F1 of \textbf{binary fact verification} on the evaluation set for all datasets in a zero-shot generalization setup. Rows correspond to the training dataset and columns to the evaluated dataset. }
\label{tbl:zero-shot-binary-results}%
\end{table*}%

\paragraph{Do artificial claims and natural claims generalize to each other?}
The bottom left of Table~\ref{tbl:zero-shot-three-class-results} shows that the model trained on natural claims generalizes badly to datasets with artificial claims, with an average F1 drop of 72\% relative to the in-domain performance on the three artificial datasets. In contrast, 
with natural claims\footnote{For fair comparison, we don't count the dataset pairs with the same data source, \textit{e.g.}, (SciFact-sent, SciFact-para)}, the model generalizes better, with an average F1 drop of 56\%  (bottom right). 
This observation supports the argument that artificial claims and natural claims have substantial differences, \textit{e.g.}, Wikipedia vs. real-world domains, high vs. less lexical overlap, and simple vs. diverse reasoning types, as discussed in \S~\ref{sec:data_processing} and related works~\cite{DBLP:conf/emnlp/WaddenLLWZCH20,DBLP:conf/acl/SaakyanCM20}. 

However, a surprising and counter-intuitive observation is that the model trained on artificial claims generalizes quite well to natural claims. As shown by the top right section, the average F1 drop narrows to 36.9\% when generalizing from artificial to natural claims, markedly better than when generalizing between natural claim datasets (56\% average drop). In particular, when trained on \textit{FEVER-sent}, the model achieves the best generalization results on 3 out of 5 datasets with natural claims. 
However, we will show in the following that the large size of artificial claims contributes a lot to its good generalization performance. 

\paragraph{Does generalization improve with more data?}

To examine whether good generalization on the FEVER and VitaminC datasets comes from their large dataset size, we conducted an experiment controlling for data size. Here, we only take 800 examples for each dataset to train the model. We show the zero-shot generalization results between the five datasets with sentence-level evidence in Table~\ref{tbl:zero-shot-three-class-800-samples-sent}. The results on all datasets are shown in Table~\ref{tbl:zero-shot-three-class-800-samples} in Appendix~\ref{sec:control-size-full}. 

We find that the model trained on natural claim datasets (Group I) can generalize to other natural claims slightly better than the model trained on artificial claim datasets (Group II) in this controlled setting. This confirms that dataset size contributes a lot to generalization ability. Tables~\ref{tbl:zero-shot-three-class-results} and~\ref{tbl:zero-shot-three-class-800-samples-sent} together show that Wikipedia-based artificial claims still generalize well to natural claims in real-world domains with the growth of dataset size, although crowd-sourced claims have been criticized to have strong annotation bias and cannot represent real-life misinformation~\cite{DBLP:conf/emnlp/SchusterSYFSB19}. 

\begin{table}[!t]
\setlength\tabcolsep{3pt}
\centering
\resizebox{0.48\textwidth}{!}{%
\begin{tabular}{l|cc|ccc}
\toprule
Train$\downarrow$ Test$\rightarrow$ & \tabincell{c}{FEVER\\-sent} & \tabincell{c}{Vita\\minC} &  \tabincell{c}{C-FEVER\\-sent} &  \tabincell{c}{SciFact\\-sent} & \tabincell{c}{Pub\\Health} \\
\midrule
FEVER-sent & --- & 22.20 & 13.66 & 19.64 & 24.57 \\
VitaminC & 16.93 & --- & 13.78 & 20.04 & \textbf{24.98} \\
\midrule
C-FEVER-sent & 16.63 & 8.36 & --- & 17.24 & 2.51 \\
SciFact-sent & 27.43 & \textbf{26.80} & \textbf{30.50} & --- & 13.06 \\
PubHealth & \textbf{28.60} & 26.69 & 18.04 & \textbf{22.22} & --- \\
\bottomrule
\end{tabular}%
}
\caption{Macro-F1 of 3-class fact verification for all datasets with sentence-level evidence in a zero-shot generalization setup. \textit{The size of training data is controlled to 800 samples for all datasets.} }
  \label{tbl:zero-shot-three-class-800-samples-sent}%
\end{table}%

\paragraph{Which type of label is more difficult to verify?}

Table~\ref{tbl:class-wise-zero-shot-evaluation} shows the breakdown of the class-wise F1 score. For each dataset, we show the average class-wise F1 when training the model on other datasets (zero-shot) and the class-wise F1 for training on the same dataset (in-domain). The results show that the \texttt{refutes} claim has the worst prediction score (in bold) almost for all datasets, in both the zero-shot and the in-domain setting. The in-domain results are in line with the empirical observation that ~\cite{DBLP:conf/emnlp/JiangBZD0B20} it is often ambiguous to differentiate between \texttt{refutes} and \texttt{NEI} claims even for trained human annotators. This difficulty still maintains in the zero-shot setting and harms the generalization results. 

\begin{table}[!t]
\setlength\tabcolsep{3pt}
\centering
\resizebox{0.48\textwidth}{!}{%
\begin{tabular}{l|ccc|ccc}
\toprule
\multirow{2}{*}{Dataset} & \multicolumn{3}{c|}{Zero-shot} & \multicolumn{3}{c}{In-domain} \\ \cline{2-7}
 & S & R & N & S & R & N \\
\midrule
FEVER-para & 33.75 & \textbf{25.85} & 34.41 & 87.05 & 85.89 & \textbf{83.81} \\
FEVER-sent & 42.52 & \textbf{28.24} & 44.38 & 91.32 & \textbf{89.42} & 87.10 \\
VitaminC & 50.06 & \textbf{20.44} & 25.96 & 94.44 & 89.42 & \textbf{76.42} \\
\midrule
C-FEVER-para & 34.14 & \textbf{24.09} & 47.76 & 68.50 & \textbf{13.56} & 51.76 \\
C-FEVER-sent & 28.29 & \textbf{19.72} & 64.00 & 62.27 & \textbf{46.40} & 78.96 \\
SciFact-para & 34.33 & \textbf{15.35} & 51.60 & 59.29 & \textbf{23.02} & 74.44 \\
SciFact-sent & 47.43 & \textbf{22.23} & 55.70 & 61.87 & \textbf{18.49} & 82.45 \\
PubHealth & 20.96 & \textbf{4.54} & 7.79 & 91.07 & 82.00 & \textbf{43.24} \\
\bottomrule
\end{tabular}%
}
\caption{Class-wise F1 of 3-class fact verification for the zero-shot generalization setup (left) and the in-domain training setup (right). S: supports; R: refutes; N: NEI. }
  \label{tbl:class-wise-zero-shot-evaluation}
\end{table}%

\paragraph{What is the impact of evidence length?}
From Table~\ref{tbl:zero-shot-three-class-results}, we find that fact verification in a dataset with document-level evidence is more difficult than in the same dataset with sentence-level evidence (an average of 13.29\% drop of in-domain F1). This is understandable since document-level evidence requires the model to additionally filter out irrelevant information. Climate-FEVER suffers the largest F1 drop of 31.86\%, compared with the slight performance drop on FEVER (4.3\%) and SciFact (3.72\%). 
A possible reason is that Climate-FEVER's document-level evidence consists of different (even contradictory) evidence sentences, which requires the model to reason over multiple sentences instead of just selecting the most relevant one. 

In terms of generalization, the datasets with sentence-level evidence in general achieve better generalization results than other datasets compared to their doc-level versions. For example, C-FEVER-sent generalizes better than C-FEVER-para on 5 of the 6 datasets excluding themselves. Models trained on sentence-level datasets generalize well to other document-level datasets, but the converse is not true. These results indicate that training the FV model on more fine-grained evidence yields better generalization. This is consistent with the intuition that providing fine-grained evidence eases models' learning in FV, showing the importance of accurate evidence retrieval. 

\subsection{Zero-shot generalization for binary FV}
Many works~\cite{DBLP:conf/emnlp/JiangBZD0B20,DBLP:conf/acl/SaakyanCM20} do not consider \texttt{NEI} claims due to their ambiguity. To explore whether our previous observations also hold for the task of \textit{binary fact verification}, we evaluate the generalization results for all 11 datasets using only the \texttt{supports} and \texttt{refutes} claims for training and evaluation, shown in Table~\ref{tbl:zero-shot-binary-results}. 
In this setting, artificial claims also generalize well to natural claims in other domains. In 6 of the 7 datasets with natural claims, the best generalization score is from a model trained on artificial claims. This also holds for the evidence length: datasets with sentence-level evidence tend to generalize better than document-level datasets. Finally, compared with the three-class result in Table~\ref{tbl:zero-shot-three-class-results}, generalization improves a lot on Climate-FEVER, SciFact, and PubHealth. The reason is that the model struggles in distinguishing between \texttt{refutes} and \texttt{NEI} claims in these datasets, as reflected by Table~\ref{tbl:class-wise-zero-shot-evaluation}. Therefore, they benefit a lot from removing the \texttt{NEI} label. 

\begin{table}[!t]
\setlength\tabcolsep{3pt}
\centering
\resizebox{0.48\textwidth}{!}{%
\begin{tabular}{lccccc}
\toprule
Train$\downarrow$ Test$\rightarrow$ & \tabincell{c}{C-fever\\-para} & \tabincell{c}{C-fever\\-sent} &  \tabincell{c}{SciFact\\-para} &  \tabincell{c}{SciFact\\-sent} & \tabincell{c}{Pub\\Health} \\
\midrule
FEVER-para & 50.04 & 45.99 & 59.91 & 68.18 & 42.81 \\
FEVER-sent & \textbf{55.13} & 51.84 & \textbf{66.12} & \textbf{76.39} & 40.90 \\
VitaminC & 50.41 & 49.80 & 58.27 & 68.59 & 37.84 \\
\midrule
SELF-few-shot & 22.74 & 10.75 & 17.24 & 33.38 & 43.62 \\
SELF-full & 44.61 & \textbf{62.54} & 52.25 & 54.27 & \textbf{72.10} \\
\bottomrule
\end{tabular}%
}
\caption{Macro-F1 of three-class fact verification for all datasets in a \textbf{few-shot generalization setup}. 
}
  \label{tbl:few-shot-three-class}%
\end{table}%

\begin{table*}[!t]
\setlength\tabcolsep{4pt}
\centering
\resizebox{0.9\textwidth}{!}{%
\begin{tabular}{cl|ccc|ccccc}
\toprule
Model & Train$\downarrow$ Test$\rightarrow$ & \tabincell{c}{FEVER\\-para} & \tabincell{c}{FEVER\\-sent} & VitaminC & \tabincell{c}{C-FEVER\\-para} & \tabincell{c}{C-FEVER\\-sent} & \tabincell{c}{SciFact\\-para} & \tabincell{c}{SciFact\\-sent} & PubHealth \\
\midrule
\multirow{3}{*}{BERT} & FEVER-para & --- & 64.04 & 33.82 & 18.15 & 29.71 & 18.53 & 18.19 & 3.20 \\
& FEVER-sent & \textbf{66.97} & --- & \textbf{54.75} & 35.39 & 26.49 & 39.27 & 39.72 & 25.95 \\
& VitaminC & 54.12 & 63.28 & --- & 39.57 & 34.93 & 40.80 & 45.51 & 22.21 \\
\midrule
\multirow{3}{*}{BioBERT} & FEVER-para & --- & 67.89 & 42.41 & 24.22 & 38.94 & 37.69 & 35.85 & 8.24 \\
& FEVER-sent & 57.18 & --- & 51.95 & 40.58 & 39.01 & 36.83 & 38.36 & 37.61 \\
& VitaminC & 51.03 & 60.34 & --- & \textbf{40.60} & \textbf{39.72} & 43.38 & \textbf{50.71} & 19.44 \\
\midrule
\multirow{3}{*}{SciBERT} & FEVER-para & --- & \textbf{68.49} & 39.73 & 20.43 & 33.84 & 28.90 & 35.53 & 6.37 \\
& FEVER-sent & 52.95 & --- & 51.84 & 35.50 & 35.68 & 34.24 & 39.46 & \textbf{36.46} \\
& VitaminC & 50.20 & 58.74 & --- & 37.99 & 38.79 & \textbf{43.55} & 45.69 & 20.66 \\ 
\bottomrule
\end{tabular}%
}
\caption{Zero-shot generalization performance (macro-F1) when \textbf{initialized with different pretraining models}.
}
\label{tbl:zero-shot-specialized-domain-results}
\end{table*}

\subsection{Few-shot generalization results}

We now consider the few-shot generalization setting, assuming access to a small number of examples from a target dataset (50 for each class in our experiment). We pre-train a model on a source dataset and then fine-tune it on the target dataset. Our goal is to analyze whether pre-training improves performance compared to training on the target alone. 

Table~\ref{tbl:few-shot-three-class} shows the macro-F1 on the evaluation set of all datasets. The rows ``SELF-few-shot'' and ``SELF-full'' show the performance of direct training on the 150 samples of the target dataset and the full target training set, respectively (without pre-training on the source dataset). Generally, pre-training on a source FV dataset and fine-tuning to the target outperform ``SELF-few-shot'' on all 5 datasets and ``SELF-full'' on 3 out of 5 datasets. This shows that pre-training on a related FV dataset helps to reduce the demand for human-annotated training data in the target domain. 

Second, \textit{FEVER-sent} obtains good generalization performance in all evaluation datasets. This strengthens our finding in Section~\ref{sec:zero-shot-generalization} that FEVER generalizes well to datasets with natural claims in real-world domains. Last, after finetuning, we see a dramatic improvement in performance compared to Table~\ref{tbl:zero-shot-three-class-results}. This highlights that current models over-fit the data they are trained on, and small amounts of data from the target distribution can overcome this generalization gap. 

\section{Improving Generalization}

We then investigate two ways to improve the generalization ability of fact verification: 1) incorporating domain knowledge via pretraining on specialized domains, and 2) automatically generating training data via data augmentation. 


\subsection{Pretraining on Specialized Domains}

In-domain knowledge is essential for fact-checking in specialized domains. For example, virology background knowledge is required to verify scientific claims regarding COVID19~\cite{DBLP:conf/emnlp/WaddenLLWZCH20}. When generalizing an FV model from one domain to another, how to endow the model with such in-domain knowledge is a challenging subject worthy of long-term study. Here we explore one simple solution: initializing the model with language models pretrained on specialized domains. 

In Table~\ref{tbl:zero-shot-specialized-domain-results}, we show the zero-shot generalization performance when initializing the FV model with BioBERT~\cite{DBLP:journals/bioinformatics/LeeYKKKSK20} (pretrained on biology literature) and SciBERT~\cite{DBLP:conf/emnlp/BeltagyLC19} (pretrained on scholarly documents). Our goal is to explore whether pretraining on specialized domains helps the generalization. To eliminate the impact of other factors such as the model size, we use the BERT model~\cite{DBLP:conf/naacl/DevlinCLT19} as the baseline, since BioBERT and SciBERT are both based on the BERT model. 

We find that BioBERT and SciBERT both outperform the BERT on the generalization scores in Climate-FEVER, SciFact, and PubHealth, with an average improvement of 21.39\% and 12.69\% in F1, respectively. However, their performance on Wikipedia-based datasets (FEVER and VitaminC) is relatively worse with BERT (-2.6\% and -17.7\% for BioBERT and SciBERT, respectively). This confirms the generalization of FV in certain domains (\textit{e.g.}, science) can be improved with the language models pretrained on relevant domains (\textit{e.g.}, scientific papers). We have similar observations for the few-shot generalization setting, shown in Table~\ref{tbl:few-shot-specialized-domain-results} in Appendix~\ref{sec:full_few_shot_results}. Despite the positive results, a suitable pretraining model in certain domains (\textit{e.g.}, tweets) is often unavailable. Moreover, this requires re-training the FV model during domain transfer. Therefore, how to develop a more accessible and less expensive way to incorporate in-domain knowledge required for fact-checking still requires further investigation. 

\subsection{Data Augmentation}
Another direction we explore is improving generalization via data augmentation, which has recently shown promising results in other NLP tasks such as question answering~\cite{DBLP:conf/emnlp/YueKF21} and machine translation~\cite{DBLP:conf/acl/ChengJME20}. 
We first train a \textit{claim generation} model based on the BART~\cite{DBLP:conf/acl/LewisLGGMLSZ20}, using the (evidence, claim, label) triples \textit{in the source domain} as training data. We use the format \texttt{[LABEL] \textit{label} [NER] \textit{NERs} [EVIDENCE] \textit{evidence}} as the input, and use the \texttt{\textit{claim}} as the target output for training, where \texttt{\textit{NERs}} are the entities appearing in the claim (we add \texttt{\textit{NERs}} to guide the model to generate more specific claims). We then apply the trained model to generate claims with different labels in the target domain by separately assigning \texttt{supports}, \texttt{refutes}, \texttt{NEI} as the label prefix of the evidence, and we randomly assign an entity from the evidence as the \texttt{\textit{NERs}}\footnote{Since no ground-truth claim is available for the target domain, the entity cannot come from the claim.} to guide the claim generation. We name this method as \textbf{BART-gen}. 

We train BART-gen on the \textit{FEVER-sent} dataset and generate synthetic training (evidence, claim, label) triples for other datasets. For each piece of evidence in the target domain, we generate six claims with different (\texttt{\textit{label}}, \texttt{\textit{NERs}}) combinations. Table~\ref{tbl:data-augmentation-three-class} shows the zero- and few-shot generalization results for BART-gen and other baselines: 1) \textit{FEVER-full}: the model is trained on the original FEVER-sent dataset; 2) \textit{FEVER-control}: the model is trained on a random subset of FEVER-sent which has the same amount of data with the generated data; 3) \textit{BART-gen}: the model is trained on the generated data. For the few-shot setting, the model is further fine-tuned with 150 in-domain samples. 

\begin{table}[!t]
\setlength\tabcolsep{3pt}
\centering
\resizebox{0.48\textwidth}{!}{
\begin{tabular}{lccccc}
\toprule
Method$\downarrow$ Test$\rightarrow$ & \tabincell{c}{C-fever\\-para} & \tabincell{c}{C-fever\\-sent} &  \tabincell{c}{SciFact\\-para} &  \tabincell{c}{SciFact\\-sent} & \tabincell{c}{Pub\\Health} \\
\midrule
\multicolumn{5}{l}{\textbf{Zero-shot setting}} \\
FEVER-full & 44.98 & 48.70 & 44.98 & \textbf{56.15} & 21.61 \\
FEVER-control & 37.14 & 45.64 & 32.42 & 47.57 & 20.69 \\
BART-gen & \textbf{46.51} & \textbf{51.67} & \textbf{47.80} & 54.63 & \textbf{69.82} \\
\midrule
\multicolumn{5}{l}{\textbf{Few-shot setting}} \\
FEVER-full & \textbf{55.13} & 51.84 & \textbf{66.12} & 76.39 & 40.90 \\
FEVER-control & 37.17 & 48.13 & 62.72 & \textbf{77.41} & 47.03 \\
BART-gen & 46.94 & \textbf{52.80} & 50.10 & 59.00 & \textbf{70.45} \\
\bottomrule
\end{tabular}%
}
\caption{Macro-F1 of three-class fact verification \textbf{with data augmentation} (BART-gen) and other baselines. The (top, bottom) shows results for (zero-, few-)shot generalization, respectively. 
}
\label{tbl:data-augmentation-three-class}%
\end{table}%

For the zero-shot setting results in Table~\ref{tbl:data-augmentation-three-class}, \textit{BART-gen} consistently improves the generalization performance compared with \textit{FEVER-full} (+24.9\% in average) and \textit{FEVER-control} (+47.4\% in average). The results show that training with generated target data is in general more effective than directly generalizing a model trained on the source data. This is better reflected by comparing \textit{BART-gen} with \textit{FEVER-control} in which the data amount is the same. The improvement is especially noticeable for \textit{PubHealth}, probably because it lacks the \texttt{NEI} claims in its original training set. Data augmentation addresses this by generating a sufficiently balanced number of claims for each label. 

However, our human evaluation in Appendix~\ref{sec:human-eval-gen-claims} shows that around 30\% of generated claims suffer the \textit{label inconsistency} problem, \textit{i.e.}, the BART-gen often generates a fluent claim that does not match our desired label (for example, we want to generate a \texttt{refutes} claim, but the generated claim is actually \texttt{NEI}). Label inconsistency may introduce conflicting information between the pretraining and fine-tuning stages, which we hypothesize is the cause for the lower level of improvement in fine-tuning the model on the generated data, compared with fine-tuning the model on FEVER. Therefore, although data augmentation is a promising direction to improve generalization, it remains a challenging problem regarding how to generate high-quality claims with consistent labels. 

\section{Related Work}


To overcome the proliferation of misinformation, a great amount of progress has been made in the area of automated fact verification. For modeling, pretraining-based models~\cite{DBLP:conf/emnlp/NieWB19,stammbach2019team,DBLP:conf/iclr/ZhaoXRSBT20,DBLP:conf/ecir/SoleimaniMW20} have been used for better text representation and have achieved promising performance. Graph-based models~\cite{DBLP:conf/acl/ZhouHYLWLS19,DBLP:conf/acl/LiuXSL20,DBLP:conf/acl/ZhongXTXDZWY20} are used to facilitate the reasoning over multiple pieces of evidence. However, most existing models rely on large-scale in-domain training data, which is often unrealistic for every domain that demand fact checking. In this paper, we aim to address this by working towards a generalizable fact verification system that can adapt to different domains with zero or few samples in the target domain. 

For datasets, various fact-checking datasets representing different real-world domains are proposed, including both naturally occurring~\cite{DBLP:conf/emnlp/AugensteinLWLHH19,DBLP:conf/acl/GuptaS20,DBLP:conf/acl/SaakyanCM20,DBLP:SciTab} and human-crafted~\cite{DBLP:conf/naacl/ThorneVCM18,DBLP:conf/lrec/SatheALPP20,DBLP:conf/naacl/SchusterFB21,DBLP:journals/tacl/AtanasovaSLA22} fact-checking claims. While these FV datasets focus on different domains, there is still a substantial overlap in the abilities required to verify claims across these datasets. However, little analysis has been done on whether they generalize to one another, and the extent to which existing datasets can be leveraged for improving performance on new ones. Similar studies have been done in other NLP tasks~\cite{DBLP:conf/acl/TalmorB19,DBLP:conf/emnlp/HardalovANA21}, while it is less investigated in fact verification. In this paper, we bridge this gap by conducting a comprehensive study of generalization and transfer across existing FV datasets, revealing several key factors for better generalization. 


\section{Conclusion and Future Work}

In this work, we perform a thorough empirical investigation of zero- and few-shot generalization over 11 fact verification datasets. Moreover, we conduct an exhaustive analysis and highlight the most important factors influencing the generalization performance. We further empirically explore two ways to improve generalization in fact verification. We highlight several practical takeaways: 

\noindent $\bullet$ Overall, the FV model generalizes poorly to unseen datasets compared with in-domain evaluation. However, performance is largely improved by fine-tuning on the target data. 

\noindent $\bullet$ Artificial claims can also generalize well to natural claims with an increase of dataset size. 

\noindent $\bullet$ Model trained on sentence-level evidence generalize better than document-level evidence. 

\noindent $\bullet$ The \texttt{refutes} claims are the most difficult to verify among the three labels. 


\noindent $\bullet$ Domain-specific pretraining and data augmentation consistently improves generalization performance, but they also leave unsolved challenges. 

In future work, we plan to experiment with more datasets, including non-English ones. We will also explore the generalization of other sub-tasks in fact-checking, \textit{e.g.}, claim detection, evidence retrieval. 

\section*{Limitations}
In our study, we primarily focused on assessing the generalization capabilities of Transformer-based models, such as RoBERTa. However, we did not extend our evaluation to include zero- and few-shot learning performance on large language models (LLMs) like InstructGPT~\cite{DBLP:InstructGPT} and GPT-4~\cite{DBLP:GPT4}, due to the high experimental costs. Recently, these LLMs have demonstrated impressive few-shot learning capacities across a variety of natural language processing tasks, including few-shot fact-checking~\cite{DBLP:conf/acl/PanWLLWKN23}. However, they are API-based and function as black-box models. This restricts our ability to delve deeper into their behavior, given that we cannot access their model weights or fine-tune them directly. On the contrary, Transformer-based models are open-sourced and replicable, providing a wealth of opportunity for more profound insights into our study and paving the way for future research. 

\section*{Acknowledgements}
This research / project is supported by the Ministry of Education, Singapore, under its MOE AcRF TIER 3 Grant (MOE-MOET32022-0001). Any opinions, findings and conclusions or recommendations expressed in this material are those of the author(s) and do not reflect the views of the Ministry of Education, Singapore. 

\bibliography{anthology,custom}
\bibliographystyle{acl_natbib}

\clearpage

\begin{table*}[th]
\setlength\tabcolsep{3pt}
  \centering
\resizebox{\textwidth}{!}{%
\begin{tabular}{lccccccc}
\toprule
Dataset & Domain & Claim & \tabincell{c}{Doc-level \\evidence} & \tabincell{c}{Sent-level\\evidence} & \tabincell{c}{NEI\\claims} & \tabincell{c}{Publicly\\available} \\
\midrule
FEVER~\cite{DBLP:conf/naacl/ThorneVCM18} & Wikipedia & artificial  & \Checkmark  & \Checkmark  & \Checkmark  & \Checkmark \\
WikiFactCheck~\cite{DBLP:conf/lrec/SatheALPP20} & Wikipedia & artificial & \Checkmark  & \Checkmark  &    & \Checkmark \\
HOVER~\cite{DBLP:conf/emnlp/JiangBZD0B20} & Wikipedia & artificial & \Checkmark  & \Checkmark  &    & \Checkmark \\
VitaminC~\cite{DBLP:conf/naacl/SchusterFB21}  & Wikipedia & artificial & \Checkmark  & \Checkmark  & \Checkmark  & \Checkmark \\
Fool Me Twice~\cite{DBLP:conf/naacl/EisenschlosDBBB21}  & Wikipedia & artificial & \Checkmark  & \Checkmark  &    & \Checkmark \\
CREAK~\cite{DBLP:journals/corr/abs-2109-01653}  & Commonsense & artificial & \Checkmark  &    &    & \Checkmark \\
\midrule
CreditAccess~\cite{DBLP:conf/cikm/PopatMSW16} & News & natural & \Checkmark  & \Checkmark  &    & \Checkmark \\
Emergent~\cite{DBLP:conf/naacl/FerreiraV16} & Emergent & natural & \Checkmark  & \Checkmark  & \Checkmark  & \Checkmark \\
MultiFC~\cite{DBLP:conf/emnlp/AugensteinLWLHH19} & Multiple & natural &   &    & \Checkmark  & \Checkmark \\
Snopes~\cite{DBLP:conf/conll/HanselowskiSSLG19} & News & natural & \Checkmark  & \Checkmark  & \Checkmark  &  \\
Climate-FEVER~\cite{DBLP:journals/corr/abs-2012-00614} & Climate & natural & \Checkmark  & \Checkmark  & \Checkmark  & \Checkmark \\
SciFact~\cite{DBLP:conf/emnlp/WaddenLLWZCH20} & Scientific & natural & \Checkmark  & \Checkmark  & \Checkmark  & \Checkmark \\
PubHealth~\cite{DBLP:conf/emnlp/KotonyaT20}  & Health & natural & \Checkmark  & \Checkmark  & \Checkmark  & \Checkmark \\
COVID-Fact~\cite{DBLP:conf/acl/SaakyanCM20} & Forum & natural & \Checkmark  & \Checkmark  &    & \Checkmark \\
X-Fact~\cite{DBLP:conf/acl/GuptaS20} & Multiple & natural & \Checkmark  &    & \Checkmark  & \Checkmark \\
FaVIQ~\cite{DBLP:journals/corr/abs-2107-02153} & Forum & natural & \Checkmark  & \Checkmark  &    & \Checkmark \\
\bottomrule
\end{tabular}%
}
  \caption{A list of candidate fact verification datasets.}
  \label{tab:dataset-list}%
\end{table*}%

\begin{figure}[!ht]
    \includegraphics[width=0.95\linewidth]{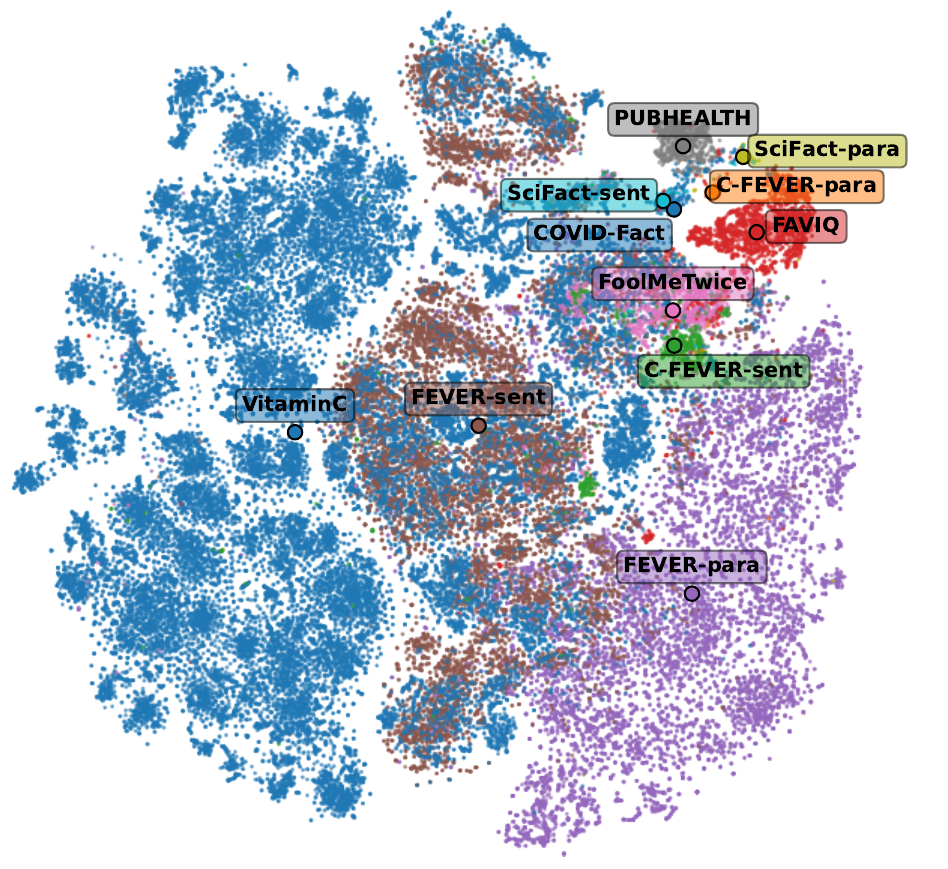}
    \caption{tSNE plot of \texttt{[CLS]} representations of each dataset; highlighted points denote cluster centroids.}\label{fig:tsne}
\end{figure}

\appendix

\section{A List of Fact Verification Datasets}
\label{sec:list-fv-datasets}
In Table~\ref{tab:dataset-list} we provide a comprehensive list of candidate datasets that we consider for our study, including those are not selected in our benchmark in the end. The candidate list does not include the fact checking datasets without providing evidence for the claim (\textit{e.g.}, FakeNewsNet~\cite{DBLP:journals/bigdata/ShuMWLL20}), or focusing on non-textual evidence such as table (\textit{e.g.}, FEVEROUS~\cite{DBLP:journals/corr/abs-2106-05707} and TabFact~\cite{DBLP:conf/iclr/ChenWCZWLZW20}). 

Afterward, we exclude some datasets from the candidate list, mainly because of the lack of clean evidence, the small scale in size, non-English claims, and unavailability. For example, we exclude Emergent~\cite{DBLP:conf/naacl/FerreiraV16} since it only contains 300 claims. We exclude X-Fact~\cite{DBLP:conf/acl/GuptaS20} since it is a multi-lingual dataset that mainly focus on non-English languages. Snopes~\cite{DBLP:conf/conll/HanselowskiSSLG19} is not included since it is not publicly available. We also exclude CREAK~\cite{DBLP:journals/corr/abs-2109-01653}, HOVER~\cite{DBLP:conf/emnlp/JiangBZD0B20}, and MultiFC~\cite{DBLP:conf/emnlp/AugensteinLWLHH19} since their evidence is either coarse-grained (\textit{e.g.}, the whole Wikipedia page) or noisy (\textit{e.g.}, the original webpage in certain fact checking website).

\section{Domain Divergence Analysis}
\label{sec:domain-divergence-analysis}
Following~\citet{DBLP:conf/emnlp/HardalovANA21}, we plot the 11 datasets in a latent vector space to visualize the global structure of the datasets. We proportionally sample 82K (10\%) examples, and we pass them through a RoBERTa-large~\cite{DBLP:journals/corr/abs-1907-11692} model without any training. The input has the following form: \texttt{[CLS] \textit{claim} [SEP] \textit{evidence}}. Next, we take the \texttt{[CLS]} token representations, and we plot them in Figure~\ref{fig:tsne} using tSNE~\cite{JMLR:v9:vandermaaten08a}.
We can see that datasets with \textit{natural claims} are grouped top-right, clearly separated from those with \textit{artificial claims}. 
The clusters of real-world domain datasets do not overlap, which highlights the rich diversity of our selected datasets.
We also notice that datasets with \textit{sentence-level} evidence have little overlap with their \textit{paragraph-level} counterparts (e.g., Climate-FEVER-sentence v.s. Climate-FEVER-paragraph). To sum up, Figure~\ref{fig:tsne} confirms that there exists divergence between different domains and datasets. 

\begin{table*}[!ht]
\setlength\tabcolsep{8pt}
\centering
\resizebox{\textwidth}{!}{%
\begin{tabular}{l|ccc|ccccc}
\toprule
Train$\downarrow$ Test$\rightarrow$ & \tabincell{c}{FEVER\\-para} & \tabincell{c}{FEVER\\-sent} & VitaminC & \tabincell{c}{C-FEVER\\-para} & \tabincell{c}{C-FEVER\\-sent} & \tabincell{c}{SciFact\\-para} & \tabincell{c}{SciFact\\-sent} & PubHealth \\
\midrule
FEVER-para & --- & \textbf{31.98} & 27.58 & 32.83 & 25.22 & \textbf{25.36} & 30.72 & 18.82 \\
FEVER-sent & 16.68 & --- & 22.20  & 21.78 & 13.66 & 20.01 & 19.64 & 24.57 \\
VitaminC & 16.70  & 16.93 & --- & 22.01 & 13.78 & 20.04 & 20.04 & \textbf{24.98} \\
\midrule
C-FEVER-para & 17.15 & 18.00 & 27.51 & --- & \textbf{31.44} & 17.24 & 18.19 & 5.04 \\
C-FEVER-sent & 16.63 & 16.63 & 8.36  & 17.51 & --- & 17.24 & 17.24 & 2.51 \\
SciFact-para & \textbf{30.66} & 30.66 & \textbf{28.30} & \textbf{33.46} & 31.36 & --- & \textbf{42.40} & 12.15 \\
SciFact-sent & 28.37 & 27.43 & 26.80  & 27.93 & 30.50  & 24.51 & --- & 13.06 \\
PUBHEALTH & 26.77 & 28.60  & 26.69 & 28.25 & 18.04 & 22.80  & 22.22 & --- \\
\midrule
SELF & 32.35 & 16.68 & 29.53 & 36.12 & 26.14 & 53.33 & 50.72 & 52.05 \\
\bottomrule
\end{tabular}%
}
\caption{Macro-F1 of \textbf{3-class fact verification} on the evaluation set for all datasets in a zero-shot generalization setup. The size of training data is controlled to 800 samples for all datasets. Rows correspond to the training dataset and columns to the evaluated dataset. The row SELF corresponds to the in-domain performance (training and testing on the same target dataset). }
  \label{tbl:zero-shot-three-class-800-samples}%
\end{table*}%

\section{Full Results of Controlled Size Generalization}
\label{sec:control-size-full}
Table~\ref{tbl:zero-shot-three-class-800-samples} shows the full results of the controlled experiment in Section~\ref{sec:zero-shot-generalization} where we only take 800 examples for each dataset to train the model. 
We find that the model trained on artificial claim datasets generalize slightly worse to natural claims compared with the model trained on artificial claim datasets in the controlled size setting. Compared with the good generalization results from artificial claims to natural claims in Table~\ref{tbl:zero-shot-three-class-results}, it shows that the size of the source dataset contributes a lot to generalization ability of fact verification. 

\begin{table*}[!ht]
\setlength\tabcolsep{5pt}
\centering
\resizebox{0.8\textwidth}{!}{%
\begin{tabular}{cl|ccccc}
\toprule
Model & Train$\downarrow$ Test$\rightarrow$ & \tabincell{c}{C-FEVER\\-para} & \tabincell{c}{C-FEVER\\-sent} & \tabincell{c}{SciFact\\-para} & \tabincell{c}{SciFact\\-sent} & PubHealth \\
\midrule
\multirow{3}{*}{BERT} & FEVER-para & 41.24 & 42.68 & 42.74 & 43.42 & 15.57 \\
& FEVER-sent & 43.84 & \textbf{45.09} & 50.23 & 55.65 & 33.45 \\
& VitaminC & 45.94 & 43.38 & 57.25 & 58.12 & 33.69 \\
\midrule
\multirow{3}{*}{BioBERT} & FEVER-para & 46.22 & 43.06 & \textbf{61.45} & 59.19 & 33.47 \\
& FEVER-sent & 47.64 & 43.43 & 48.59 & 53.17 & 39.44 \\
& VitaminC & 44.16 & 40.48 & 54.52 & \textbf{61.62} & 32.29 \\
\midrule
\multirow{3}{*}{SciBERT} & FEVER-para & \textbf{47.92} & 40.46 & 53.12 & 56.83 & 34.39 \\
& FEVER-sent & 43.92 & 40.94 & 56.53 & 60.49 & \textbf{42.14} \\
& VitaminC & 46.23 & 40.16 & 60.95 & 61.04 & 37.37 \\
\bottomrule
\end{tabular}%
}
\caption{Few-shot generalization performance (macro-F1) when \textbf{initialized with different pretraining models}.
}
\label{tbl:few-shot-specialized-domain-results}
\end{table*}

\section{Full Results of Few-shot Generalization}
\label{sec:full_few_shot_results}
In Table~\ref{tbl:few-shot-specialized-domain-results}, we show the few-shot generalization performance of FV model pretrained on specialized domains.
After finetuning, we observe dramatic improvement in performance comparing to Table~\ref{tbl:zero-shot-specialized-domain-results} (+14.31\% for BERT, +11.84\% for BioBERT, +15.29\% for SciBERT). Under few-shot setting, we find that BioBERT and SciBERT still outperform the BERT on the generalization scores in all datasets except Climate-FEVER-sentence.

\section{Human Evaluation of Generated Claims}
\label{sec:human-eval-gen-claims}

We conduct the human evaluation on the claims generated by \textit{BART-gen} on four datasets: Climate-FEVER-sentence, Sci-Fact-sentence, PubHealth, and COVID-Fact. We randomly sample $90$ generated claims for each dataset with a balanced desired label distribution. To be specific, 30/30/30 of their \textit{desired labels}, \textit{i.e.}, the type of claim we expect the model to generate (by appending the corresponding label prefix) are \texttt{supports}/\texttt{refutes}/\texttt{NEI}. We ask two expert human annotators to annotate the \textit{actual label} for each claim, \textit{i.e.}, whether the generated claim is supported, refuted, or cannot be verified by the evidence. If the generated claim is an incomplete or unreadable sentence, we label it as \texttt{Unclassified}.

\begin{figure*}[!ht]
     \centering
     \begin{subfigure}[b]{0.49\textwidth}
         \centering
         \includegraphics[width=\textwidth]{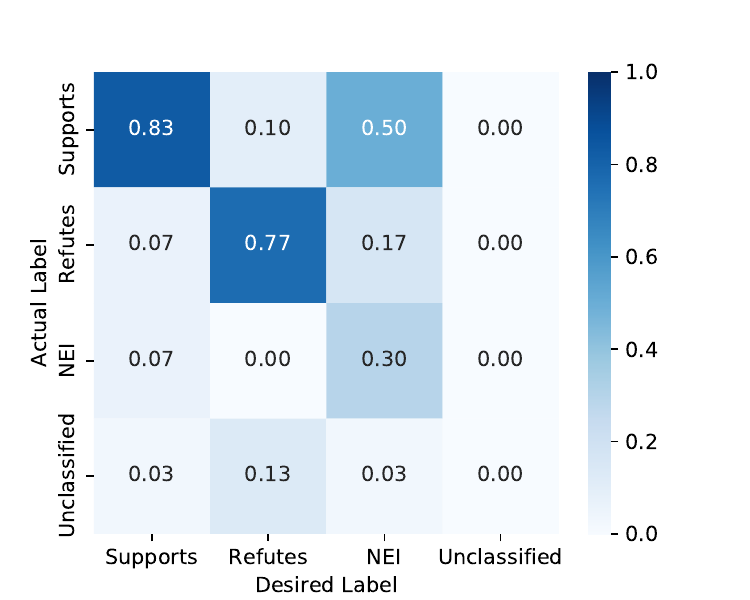}
         \caption{Climate-FEVER-sentence}
         \label{fig:climate-fever-cf_matrix}
     \end{subfigure}
     \hfill
     \begin{subfigure}[b]{0.49\textwidth}
         \centering
         \includegraphics[width=\textwidth]{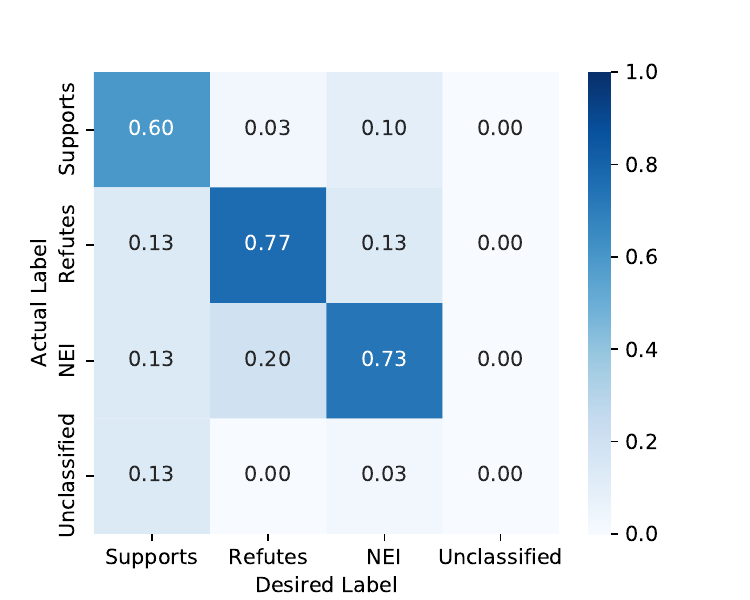}
         \caption{SciFact-sentence}
         \label{fig:scifact-sent-cf_matrix}
     \end{subfigure}
     \hfill
     \begin{subfigure}[b]{0.49\textwidth}
         \centering
         \includegraphics[width=\textwidth]{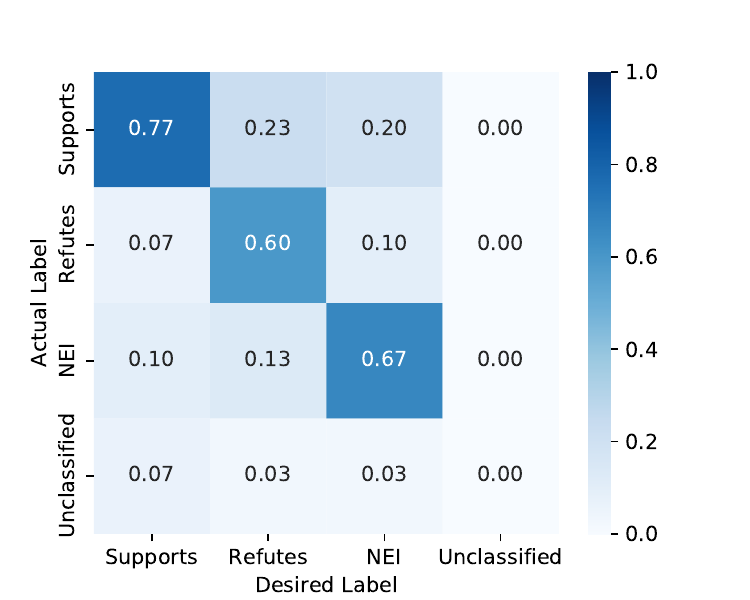}
         \caption{PubHealth}
         \label{fig:pubhealth-cf_matrix}
     \end{subfigure}
     \hfill
     \begin{subfigure}[b]{0.49\textwidth}
         \centering
         \includegraphics[width=\textwidth]{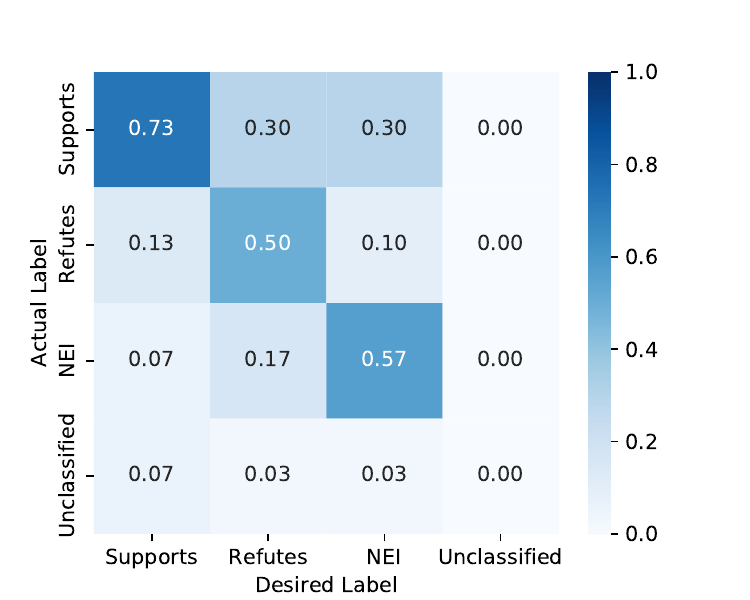}
         \caption{COVID-Fact}
         \label{fig:covid-fact-cf_matrix}
     \end{subfigure}
        \caption{Confusion matrices (normalized over columns) of generated claims on four datasets. The desired label is the input label for our claim generation model (BART-gen) and the actual label is the human-annotated label. ``Unclassified'' means that the generated claim is incomplete or unreadable.}
        \label{fig:human_evaluation_claims}
\end{figure*}

Figure~\ref{fig:human_evaluation_claims} shows the confusion matrix for the desired labels and the actual labels. We find that in all four datasets, around 30\% of the generated claims suffer from the \textit{label inconsistency} problem, \textit{i.e.}, the actual label of the claim is not the desired label. Specially, the confusion between the \texttt{refutes} and \texttt{NEI} claim is the major type of error, showing that \texttt{refutes} and \texttt{NEI} claims are the hardest for the model to generate. 

We also observe that around 5\% of the generated claims are incomplete or unreadable. Moreover, most generated claims are short and simple (e.g., ``\textit{Gilbert Rothschild was a person}''), which do not require complex reasoning to verify. It is therefore worthy to investigate how to obtain high-quality claims in data augmentation for better generalization in the future study.

\end{document}